\documentclass[sigconf]{acmart}
\AtBeginDocument{%
  }

\setcopyright{acmlicensed}
\copyrightyear{2018}
\acmYear{2018}
\acmDOI{XXXXXXX.XXXXXXX}

\acmConference[arXiv]{Make sure to enter the correct
  conference title from your rights confirmation emai}{Oct 31,
  2024}{preprint}
\acmISBN{978-1-4503-XXXX-X/18/06}

\usepackage{algorithm}
\usepackage{algorithmic}
\usepackage{bbm}
\usepackage{caption}
\usepackage{subcaption}

\newcommand{\R}{\mathcal{R}}
\newcommand{\V}{\mathcal{V}}
\newcommand{\E}{\mathcal{E}}
\newcommand{\G}{\mathcal{G}}

\newcommand{\Prob}{\mathbb{P}}
\newcommand{\Exp}{\mathbb{E}}
\newcommand{\Real}{\mathbb{R}}
\newcommand{\Ind}{\mathbbm{1}}

\newcommand{\jn}[1]{}
\renewcommand{\jn}[1]{{\color{red} JN: {#1}}}

\newcommand{\jean}[1]{}
\renewcommand{\jean}[1]{{\color{green} JL: {#1}}}

\newcommand{\eval}{IME Process}
\newcommand{\fullmethod}{Node Coherence Rate for Representation Interpretation}
\newcommand{\method}{NCI}

\newtheorem{theorem}{Theorem}[section]

\newtheorem{lemma}[theorem]{Lemma}
\newtheorem{definition}{Definition}[section]

\setcopyright{none}
\begin{document}

\title{Rethinking Node Representation Interpretation \\through Relation Coherence}

\author{Ying-Chun Lin}
\affiliation{%
  \institution{Purdue University}
  \country{}
}
\email{lin915@purdue.edu}

\author{Jennifer Neville}
\affiliation{%
  \institution{Microsoft}
  \country{}
}
\email{jenneville@microsoft.com}

\author{Cassiano Becker}
\affiliation{%
  \institution{Microsoft}
  \country{}
}
\email{casbecker@microsoft.com}

\author{Purvanshi Metha}
\affiliation{%
  \institution{Microsoft}
  \country{}
}
\email{purvanshi@lica.world }

\author{Nabiha Asghar}
\affiliation{%
  \institution{Microsoft}
  \country{}
}
\email{naasghar@microsoft.com}

\author{Vipul Agarwal}
\affiliation{%
  \institution{Microsoft}
  \country{}
}
\email{vipulag@microsoft.com}



\renewcommand{\shortauthors}{Lin et al.}

\begin{abstract}
Understanding node representations in graph-based models is crucial for uncovering biases, diagnosing errors, and building trust in model decisions. However, previous work on explainable AI for node representations has primarily emphasized {\em explanations} (reasons for model predictions) rather than {\em interpretations} (mapping representations to understandable concepts). 
Furthermore, the limited research that focuses on interpretation lacks validation, and thus the reliability of such methods is unclear. We address this gap by proposing a novel interpretation method---{\fullmethod} ({\method})---which quantifies how well different node relations are captured in node representations. We also propose a novel method (IME) to {\em evaluate}  
the accuracy of different interpretation methods.  Our experimental results demonstrate that {\method} reduces the error of the previous best approach by an average of $39\%$. We then apply {\method} to derive insights about the node representations produced by several graph-based methods and assess their quality in unsupervised settings. Our code is published on Github\footnote{https://github.com/ycjeanlin/Node-Coherence-Rate-for-Representation-Interpretation.git}.
\end{abstract}




\received{20 February 2007}
\received[revised]{12 March 2009}
\received[accepted]{5 June 2009}

\maketitle

\section{Introduction}
In graph data, nodes represent real-world entities, and their relationships are depicted by edges. Node representation learning involves transforming each node into a meaningful and low-dimension vector (embedding) within the embedding space. Identifying human-comprehensible factors in such node representations is crucial because it enables bias identification, error detection, and establishment of trust. We typically depend on two types of methods to comprehend machine learning models: interpretation and explanation. As described in~\cite{interpret_explain_dnn,trust_ml}, interpretation refers to making the output structures, e.g., node representations (embeddings), understandable for humans, while explanation makes the connections between parameters and features behind a specific model decision, such as classification or regression.

\begin{figure}[t]
\centering
\begin{subfigure}{\linewidth}
  \centering
  \includegraphics[trim={0 0 0 0},clip,width=\linewidth]{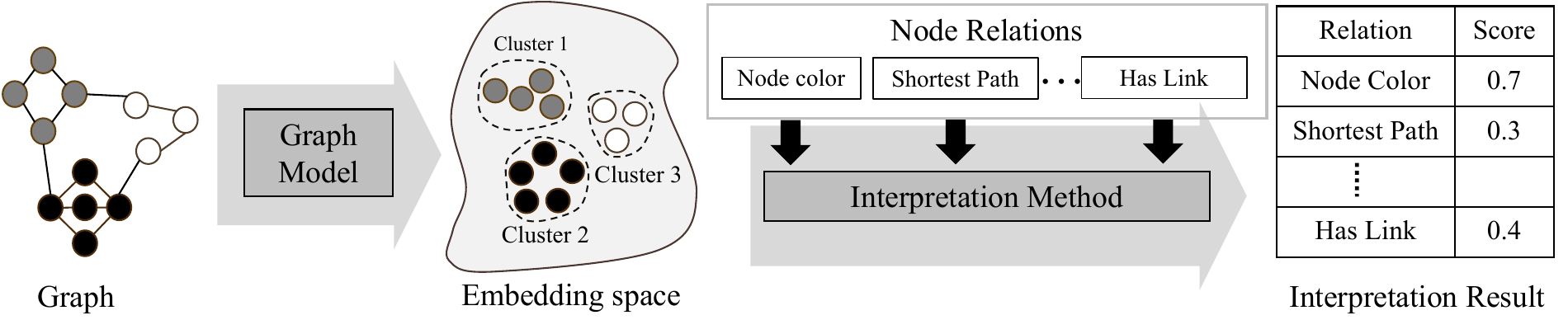}
  \caption{Representation Interpretation Process.}
  \label{fig:overview_interpret}
\end{subfigure}\\
\vspace{3mm}
\begin{subfigure}{\linewidth}
  \centering
  \includegraphics[trim={0 0 0 0},clip,width=\linewidth]{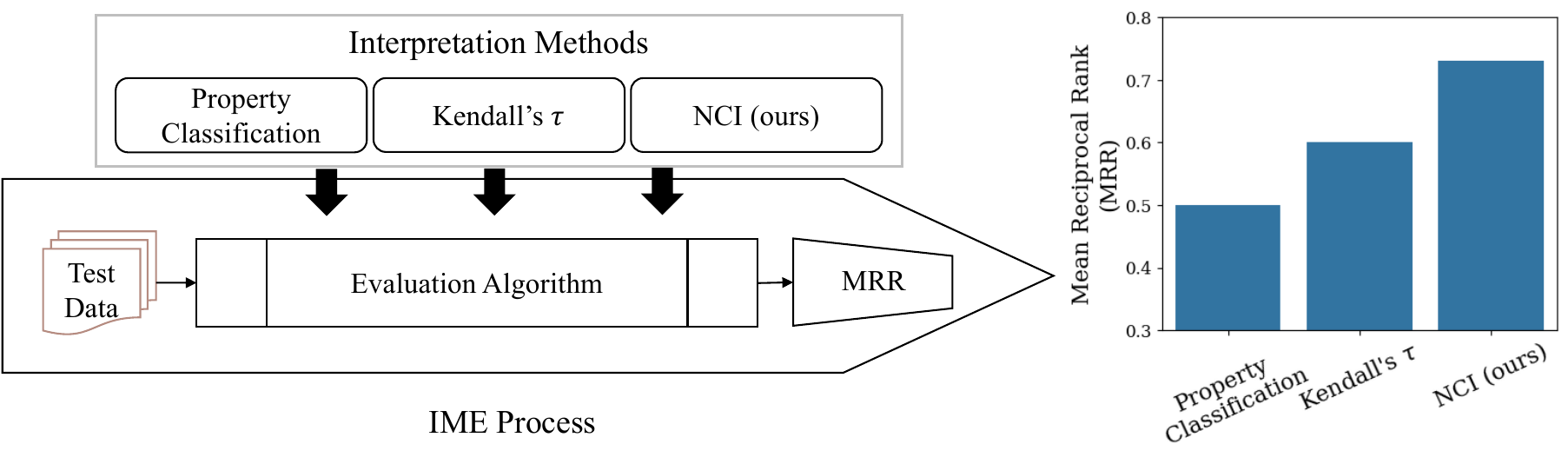}
  \caption{Interpretation Method Evaluation (IME).}
  \label{fig:overview_ime}
\end{subfigure}
\vspace{-3mm}
\caption{Illustration of representation interpretation process and Interpretation Method Evaluation (IME) process.}
\label{fig:overview}
\vspace{-5mm}
\end{figure}

Previous studies on understanding graph-based models have concentrated on explaining model decisions~\cite{gnnexplainer,graphlime,graphmask} and there has been less emphasis on interpreting learned node representations by translating  the output embeddings into human-understandable concepts. Therefore, a thorough investigation of node representations interpretation is necessary. Some interpretation approaches~\cite{eval_stuct,intrinsic_struct_eval,interpret_dimension} interpret node embeddings based on their ability to capture the underlying relations between nodes in the embedding space, as depicted in Figure~\ref{fig:overview_interpret}. However, these works lack explicit measurements of accuracy, which makes it difficult to evaluate the reliability of the proposed interpretation methods. In Section~\ref{sec:eval_process}, we will outline an  Interpretation Method Evaluation (IME) process. When we apply this evaluation to previous interpretation methods, they typically have low {\em interpretation accuracy} (more specifically Mean Reciprocal Rank). In Figure~\ref{fig:overview_ime}, up to $50\%$ of time the incorrect interpretation is made by Property Classification. See Figure~\ref{fig:ime} for an illustration of how we calculate interpretation accuracy.


In this paper, our objective is to conduct a comprehensive investigation into  node representation interpretation. Specifically, we (1) propose an algorithm (IME) to evaluate interpretation accuracy for all interpretation methods, (2) introduce a new interpretation method (NCI) that is more accurate, and (3) employ our interpretation method to understand and select graph-based models in unsupervised scenarios.

Specifically, we propose a novel interpretation method, {\fullmethod} ({\method}). This method aims to determine the node relations captured by a learned node representation by assessing their {\em relation coherence}. Relation coherence for node representations refers to the notion  that when a given node embedding captures a particular node relation, and two nodes exhibit higher similarity in terms of this relation, their embedding distance should be smaller. {\em Coherence rate} is our interpretation score, which indicates the degree to which a node relation is captured by node embeddings. Finally, we develop {\em model coherence score} for model selection based on the interpretation results, for use in unsupervised settings.

In summary, the main contributions of this work are: \\
(1) We propose the {\eval} to evaluate {\em all} interpretation methods, which  generates node representations with ground-truth node relations and tests whether an interpretation method can output the correct interpretation result. \\
(2) We develop {\method} to interpret node representations by assessing the extent to which different node relations are captured in the embedding space. We also derive two bounds to demonstrate whether a node relation is significantly captured.  \\
(3) We demonstrate the effectiveness of {\method} on six real graphs. The results demonstrate that our method is more accurate in practice, reducing interpretation errors by an average of $39\%$ compared to Kendall's based on the results in Table~\ref{tab:interpret_mrr}. \\
(4) We include interpretation results for five graph-based models, which demonstrate the characteristics of the embeddings learned by different models. We also demonstrate how {\method} can help in selecting node representations that provide higher-quality features for various downstream tasks.

\section{Preliminaries}
\label{sec:def}

{\em Unsupervised Node Representation Learning} focuses on learning meaningful and informative representations (embeddings) for nodes within a graph or network. A graph is formally defined as $\G=(\V, \E, X, Y)$, where $\V$ is the node set, $\E$ is the edge set, $X\in\Real^{|\V|\times d}$ is the node attributes, where $d$ is the dimension of the attributes. $Y\in[0,1]^{|\V|\times C}$ is the node labels, where $C$ is the number of classes. $A\in[0,1]^{|\V|\times|\V|}$ is the adjacency matrix representing edges in a graph. $A_{u,v}=1$ if $(u,v)\in\E$, otherwise $A_{u,v}=0$. Let $f$ be a graph-based model that learns node representations for nodes based on unsupervised objective functions. Typical objective functions include (1) graph reconstruction~\cite{gae}, which makes connected nodes near in the embedding space, or (2) contrastive functions~\cite{dgi}, which make nodes with the same attributes near in the embedding space and nodes with different attributes far away. The choice of objective function depends on the learning goals. The learned node representation is $Z=f(\G)$ and $Z$ can be used as features for various target tasks, such as node classification and link prediction. Specifically, a task-specific model $g$ takes the input $Z$ and generates task-specific outcomes. The training of $f$ and $g$ can be done jointly or independently, depending on the desired training targets. In this study, our main focus is on separately training $f$ and $g$ (i.e., an unsupervised setting).

\begin{definition} \textbf{Node Relation}.
Refers to a specific relationship between nodes in a graph indicating the similarity (or dissimilarity) between pairs of nodes. Let $\R=\left\{r_1, r_2, \cdots, r_{|\R|}\right\}$ be the set of node relations among nodes in a graph. For each relation $r\in\R$, a similarity measure $s_r:\V\times\V\rightarrow [0,1]$ is defined to quantify the strength of the relation between two nodes. The similarity matrix $S_r\in [0,1]^{|\V|\times|\V|}$ describes the similarity values between all pairs of nodes for node relation $r$.
\end{definition}
The node relations used in this work are defined in Section~\ref{sec:node_relation}, which includes for example Shortest Path Distance and Has Link.

\begin{definition}\textbf{Relation Coherence}\label{def:relation_coherence} Consider a node triplet $(u,v,v')\in \V^3$  s.t. $s_r(u,v)\geq s_r(u,v')$ for node relation $r$, and embedding $Z=f(\G)$ where $d_f(u,v)$ is the pairwise embedding distance in the embeddings generated by $f$.
If embedding $Z=f(\G)$ is coherent with respect relation $r$ then the following hold:
\\
1. if $s_r(u,v)-s_r(u,v')>0$ then $d_f(u,v')-d_f(u,v)>0$\\
2. if $s_r(u,v)-s_r(u,v')=0$ then $d_f(u,v')-d_f(u,v)=0.$\\
\end{definition}

Relation coherence states that if a node pair $(u,v)$ shares a higher similarity w.r.t. $r$ compared to another node pair $(u,v')$, then the embedding distance between $u$ and $v$ will be smaller than that of $u$ and $v'$. A simple metric to assess the amount of relation coherence that holds for a node embedding is the portion of triples in the associated graph for which this condition holds.

Note that we don't define closeness by requiring an absolute distance between a node pair  (e.g., $s(u,v)>\eta\rightarrow d_f(u,v)<\eta'$ where $\eta$ and $\eta'$ is some value thresholds) 
because embeddings computed for same graph might vary significantly when learned from different models. Some distance values may be small in one embedding space but large in another space. The absolute values do not matter if the relative distances still preserve relative similarities.

\begin{definition}\textbf{Interpretation Method}\label{def:interpret_method}
Refers to a method $m$ that 
assesses how much a node representation $Z$ 
{\em captures} a node relation $r$ (in $G$) through
an {\em interpretation scoring function} $\Gamma_m: (Z,\G,r)\rightarrow \Real$. A higher interpretation score indicates that $Z$ reflects the relation $r$ more in its embedding distances. 
\end{definition}
Note that different interpretation methods may have different value ranges for $\Gamma_m$. For example, methods that use Kendall's $\tau$~\cite{kandell, eval_stuct} produce values in the range $[-1, 1]$ and methods based on property classification~\cite{intrinsic_struct_eval,eval_embed1,eval_embed2,node_embed_classification} use the range $[0,1]$.

An interpretation method can assess the degree of association between $Z$ and $r$ in different ways but often the notion of $Z$ capturing $r$ is not formally operationalized. We formalize it with our notion of relation coherence and measure the {\em amount} of relation coherence as the proportion of node triplets in a graph that meet the criteria in Def.~\ref{def:relation_coherence}. 
In the rest of our discussion, we will refer to the concept of "$Z$ capturing $r$" as a shorthand for indicating that $Z$ exhibits high degree of relation coherence with respect to $r$.

\section{Interpretation Method Evaluation}
\label{sec:eval_process}
To formally evaluate the accuracy of interpretation methods, we first define interpretation accuracy, then describe evaluation process. 
We define our accuracy metric based on a comparison between  interpretation scores:

\begin{definition}\textbf{Interpretation Accuracy}\label{def:interpret_acc}
Let $m$ represent an interpretation method and $\Gamma_m$ denote its interpretation scoring function. Assume that $Z$ captures $r$ better than all $r'\in\R\setminus r$. 
If 
$$\Big|\left\{r'\big|\Gamma_m(Z,\G,r)< \Gamma_m(Z,\G,r'), r'\in\R\setminus r \right\}\Big|=0,$$
then method $m$ is considered accurate, otherwise it is considered inaccurate
because the interpretation scores fail to indicate that $Z$ captures $r$ better than all $r'\in\R\setminus r$. 
\end{definition}

The above definition indicates that an {\em accurate} interpretation method $m$ should produce an interpretation score $\Gamma_m(Z_{r},\G,r)$ that is higher than $\Gamma_m(Z_{r},\G,r')$ for all $r'\in\R\setminus r$. We operationalize 
this in our Interpretation Method Evaluation (IME) Process. The goal of our method is to synthesize a ground-truth embedding $Z_{r}$ which captures a particular relation $r$ {\em better} than all other relations $r'\in\R\setminus r$. The goal is not just to produce the embedding, but to use it as ground truth to evaluate the scores/rankings that the methods produce to assess their accuracy.


\begin{algorithm}[t]
\caption{{\eval}}\label{alg:eval_algo}
\begin{algorithmic}[1]
\REQUIRE semantic relations $\R=\{r_1, r_2, \cdots, r_{|\R|}\}$, embedding dimension $d$, interpretation scoring function $\Gamma_m$ of interpretation method $m$
\ENSURE accuracy $\frac{1}{|\R|}\sum_{i=1}^{|\R|}1/rank_i$
\STATE $h=0$
\FORALL{$i\in [1,2,\cdots,|\R|]$}
    \STATE $r_t \gets r_i$
    \STATE $U, \Sigma \gets EVD(S_{r_t})$
    \STATE $Z_{r_t}=U_{|V|\times d}\Sigma_{d\times d}^{1/2}$
    \FORALL{$j\in [1,2,\cdots,|\R|]$}
        \STATE $\kappa_{r_j}=\Gamma_m(Z_{r_t}, \G, r_j)$
    \ENDFOR
    \STATE $rank_i = \left|\left\{\kappa_j\big|\kappa_i<\kappa_j \wedge i\neq j\right\}\right|+1$
\ENDFOR
\end{algorithmic}
\end{algorithm}

The challenge is then how to generate $Z_{r}$ such that it captures $r$ better than other relation $r'\in\R\setminus r$. Specifically, we are interested in the most {\em expressive} embeddings that are coherent with $r$. We define this as follows.

\begin{definition} \textbf{Most Expressive Embeddings for $r$}.\label{def:most_expressive} 
If a node embedding $Z_{r}=f(\G|\theta^*)$ is most expressive for node relation $r$, then $Z$ satisfies relation coherence for all $(u,v,v')\in\V^3$. The most expressive embedding for $r$ is denoted as $Z^{*}_r$.
\end{definition}

Note that incorporating multiple relations into the ground truth embeddings for evaluation would complicate our assessment of interpretation accuracy, so we aim to define a process that can produce a most expressive embedding $Z^{*}_r$ for a single relation $r$. We show next that  
we can use eigenvalue decomposition (EVD) to achieve this.

\begin{lemma}\label{lm:evd_most_expressive}
Let $U\Sigma U^{\top}$ be the EVD of $S_r$, where $U$ is a square matrix containing eigenvectors and $\Sigma$ is a diagonal matrix containing eigenvalues. The embeddings can be generated based on $Z_r=U_{|V|\times d}(\Sigma_{d\times d})^{1/2}$ with dimension $d\leq|\V|$, assuming that the eigenvalues in $\Sigma$ are sorted from most to least significant ones. If the embedding dimension $d=|\V|$, then $Z^{*}_r=U\Sigma^{1/2}$ is the most expressive embeddings for node relation $r$. The proof is Appendix~\ref{sec:proof_3_1}.
\end{lemma}

Based on Lemma~\ref{lm:evd_most_expressive}, the most expressive embeddings $Z^{*}_r$ for node relation $r$ can be generated by EVD of $S_r$. We use this to define the {\eval}, which will test the accuracy of an interpretation method $m$ by generating $Z^{*}_r$ with eigenvalue decomposition and comparing $\Gamma_m(Z^{*}_r,\G, r)$ to $\Gamma_m(Z^{*}_r,\G, r')$ for any other $r' \in \R\setminus r$. If $\Gamma_m$ is accurate, it should satisfy the condition that $\Gamma_m(Z^{*}_r,\G,r)>\Gamma_m(Z^{*}_r,\G,r')\forall r'\in\R\setminus r$ because $Z^{*}_r$ is most expressive for $r$.

\begin{figure}[t]
  \centering
  \includegraphics[width=\linewidth]{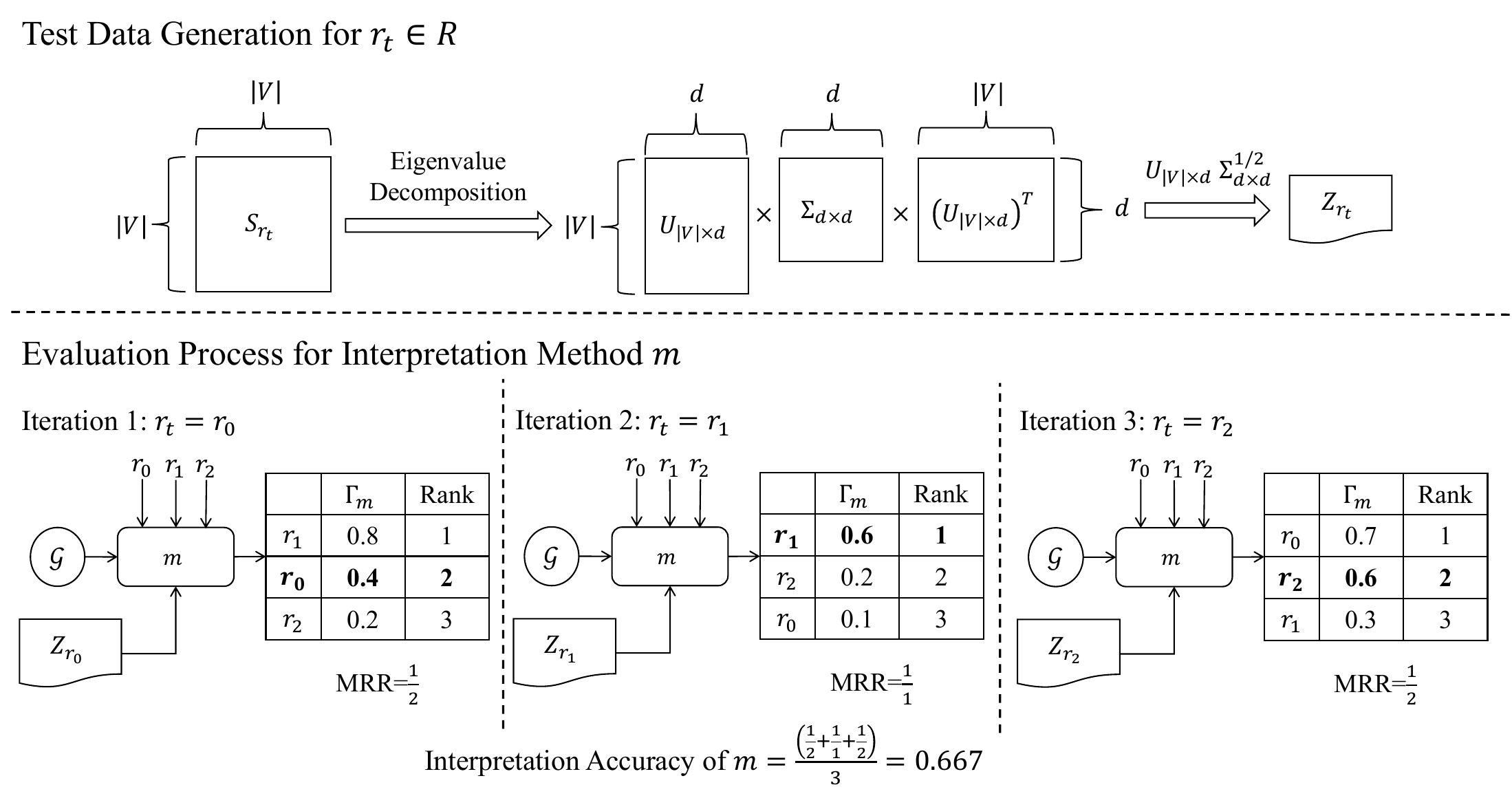}
   \caption{Illustration of IME process of evaluating the interpretation accuracy for interpretation method $m$.}
  \label{fig:ime}
  \vspace{-3mm}
\end{figure}

In addition, we can also restrict the dimension of the eigenvalue decomposition (ie. $d<|\V|$) to generate $Z_r$ to test (1) whether an interpretation method can interpret embeddings accurately in practical settings with less than fully expressive embeddings  and (2) whether the performance of each interpretation method is affected by the expressiveness of $Z_r$. This is important for more real-world settings where the goal is to learn low rank representations of nodes that are quite far from being {\em most expressive}. Figure~\ref{fig:ime} (top row) illustrates how to generate $Z_r$. 



The details of the {\eval} are in Algorithm~\ref{alg:eval_algo} and Figure~\ref{fig:ime}. At line $6$, each interpretation method $m$ uses its scoring function $\Gamma_m$ to estimate how well $Z_{r_t}$ captures each $r_i$ as $\kappa_i$. Since $Z_{r_t}$ is generated from $S_{r_t}$, it is expected that $Z_{r_t}$ captures $r_t$ well. Figure~\ref{fig:ime} (bottom row) shows an example of how to calculate the final interpretation accuracy over three relations.
If $m$ correctly interpret $Z_{r_t}$, $rank_t$ shoud be 1 (line 8). The performance metric is Mean-Reciprocal-Rank (MRR) to check $\Gamma_m(Z_{r_t},\G, r_t)>\Gamma_m(Z_{r_t},\G,r_i), \forall r_i\in\R\setminus r_t$. 

\vspace{-2mm}
\section{\fullmethod}
\label{sec:framework}
\vspace{-1mm}

Next, we propose a novel interpretation method, {\fullmethod} ({\method}) which incorporates two types of coherence that have been shown to be important for `good' node representations: {\em clustering coherence} and {\em smoothness coherence} (see e.g., \cite{representation_review}). 
Smoothness entails that, if two entities share high similarity due to a node relation, their representations should be similar (small pairwise embedding distance). Clustering refers to the fact that embeddings generally inhabit specific regions within the space, forming minimally overlapping clusters. Motivated by these notions, our {\method} interprets node representations by examining clustering and smoothness properties that are  exhibited specific to a node relation. Our overall method is depicted in Figure~\ref{fig:framework}.
{\method} scores node representations based on a relaxed notion of 
{\em weak relation coherence}. 
We will discuss the reason for using weak relation coherence and define it in the following section.

\subsection{Weak Relation Coherence}

When an embedding is less than fully expressive, it is a {\em lossy} representation, which means relation coherence will not hold for some triplets and there will be some amount of incoherence between the pairwise similarities and embedding distances. 
Specifically, if $s_r(u,v)-s_r(u,v')\leq\delta$ for some small $\delta>0$, then their embedding distance may exhibit incoherence (ie. $-\epsilon<d_f(u,v')-d_f(u,v)<0$). 
Since $d_f(u,v')-d_f(u,v)<0$, this violates relation coherence. 
In fact, any amount of incoherence will violate the conditions of Def.~\ref{def:relation_coherence}. However, if the incoherence is small (eg. for $\epsilon<0.01$), the relative ranking between node pairs will be mostly preserved and embeddings with small incoherence should be preferred to embeddings with large incoherences (ie. where relative rankings are preserved less).

We found that relation coherence is not robust to minor degradation in representation expressiveness, because small amounts of incoherence can violate the conditions of Def.~\ref{def:relation_coherence} (shown in Section~\ref{sec:exp_interpret_acc}).
To reduce the impact of limited representation expressivity in practical settings, we define a weaker notion of relation coherence:
\begin{definition} \textbf{Weak Relation Coherence}\label{def:weak_sc}. 
Consider a node triplet $(u,v,v')\in \V^3$  s.t. $s_r(u,v)\geq s_r(u,v')$ for node relation $r$, and embedding $Z=f(\G)$ where $d_f(u,v)$ is the pairwise embedding distance in the embeddings generated by $f$.
If embedding $Z=f(\G)$ is weakly coherent with respect relation $r$ then the following hold: \\
1. if $s_r(u,v)-s_r(u,v')>\delta$ then $d_f(u,v')-d_f(u,v)>0$\\
2. if $s_r(u,v)-s_r(u,v')< \delta$ then $d_f(u,v')-d_f(u,v)>-\epsilon$

\noindent for some $\delta, \epsilon>0$. 
\end{definition}

Note that this relaxes the conditions in Def.~\ref{def:relation_coherence}, requiring that the first condition only holds for triplets with larger differences in pairwise similarity. For triplets with smaller differences in pairwise similarity, there is a tolerance for the difference in embedding distances to be slightly negative. Similar to  Def.~\ref{def:relation_coherence}, the amount of weak relation coherence that holds for a node embedding can be calculated by the portion of triples for which this condition holds. We will operationalize this in Section~\ref{sec:clustering_coherence} and Section~\ref{sec:rank_coherence}.

We empirically show that interpretation scores based on weak relation coherence are more robust to lossy representation due to expressiveness reduction, and thus more accurately distinguish between the embeddings generated based on $r$ or $r'$ in our {\eval}. Therefore, we base our {\method} on two interpretation scores that correspond to notions of {\em weak relation coherence}.


\vspace{-3mm}
\subsection{Clustering Coherence}
\label{sec:clustering_coherence}

Our first approximation of weak relation coherence is {\em clustering coherence}, which requires that nodes sharing high similarity of a node relation should be closer in the embedding space (cluster), while nodes with distinct similarity should be apart from each other.

Formally, $C_{r,u}(\eta)=\{v|s_r(u,v)\geq\eta, v\in\V_t\}$ is the cluster of nodes having high similarity w.r.t $r$ given a query node $u$, nodes with low similarity values, e.g., $C'_{r,u}(\eta)=\{s_r(u,v)<\eta\}$, as {\em intruders} for $u$. The query node $u$ is drawn from a set of query nodes $\V_h\subset\V$ and the paired nodes $v$ are drawn from a target node set $\V_t\subset\V$. When the value of $\eta$ is specified, it corresponds to a value of $\delta=1-\eta$ for weak coherence in Def.~\ref{def:weak_sc}.

The clustering coherence rate is the probability that node embeddings satisfy clustering coherence. A probability measure is used to accommodate the randomness from node embeddings, which may not be most expressive to any node relations. The randomness will be discussed in Theorem~\ref{thm:ub_cc_rate}. The clustering coherence rate is:
\begin{definition} \textbf{Clustering Coherence Rate}.
For a graph-based model $f$, node relation $r$, and query node $u$, clustering coherence rate is the probability that {\em similar node} $v_s\in C_{r,u}(\eta_s)$ and {\em intruder} $v_i\in C'_{r,u}(\eta_i)$ can be distinguished by node embeddings:
\begin{align}\label{eq:cc_prob}
    \Prob \Big(d_f(u,v_i)-&d_f(u,v_s)\geq c\sigma_{D_f} \notag\\
    &\: \Big\vert \: v_s\in C_{r,u}(\eta_s) \wedge v_i\in C'_{r,u}(\eta_i)\Big)
\end{align}
Here, we define hyperparameter $c>0$ and $\sigma_{D_f}$ as the standard deviation of the distribution of pairwise distances w.r.t. $f$. Since we normalize all embeddings before calculating clustering coherence rates, the $c\sigma_{D_f}$ here ensures that $v_s\in C_{r,u}(\eta_s)$ is close to $u$. Thus, small amounts of incoherence can be ignored for nodes close to $u$. Note that we set $\eta_s>\eta_i$ to provide some separation in $r$ between $C_{r,u}(\eta_s)$ and $C'_{r,u}(\eta_i)$.

To derive clustering coherence rate, we define a function to calculate the empirical probability associated with Eq.~\eqref{eq:cc_prob} over a set of query nodes $\V_h$, as follows:
\allowdisplaybreaks{}
\begin{align}\label{eq:cc_test}
    \Gamma_{clu}(f,r,c,\eta_i,\eta_s,k)=& \notag\\
    \frac{1}{|\V_h|\times K}\sum_{u \in \V_h} \sum_{k=1}^{K}\Ind&\Big\{d_f(u,v_i)-d_f(u,v_s)\geq c\sigma_{D_f} \notag\\
    &\: \Big\vert \:(v_s,v_i)_k\in C_{r,u}(\eta_s)\times C'_{r,u}(\eta_i)\Big\}.
\end{align}
\end{definition}


How to set a reasonable margin $c\sigma_{D_f}$ in Eq.~\eqref{eq:cc_test} has to consider the variance in embeddings distances. The distance variance can impact the likelihood that a query node having clustering property just by chance. If the margin is small, many nodes can easily have clustering property such that the level of clustering coherence is mis-interpreted. To ensure a fair comparison among models, we choose a variance-based threshold such that all models have an equal clustering coherence rate under the null hypothesis that the embedding does not capture a node relation but have some level of clustering coherence by chance.

\begin{theorem}\label{thm:ub_cc_rate}
Given random two nodes $v, v'\in\V_t$, a query node $u$ and the decision margin $c\sigma_{D^f}$, the probability that $d_f(u,v)-d_f(u,v')\geq c\sigma_{D_f}$ even if their embedding distance does not affected by the node relation $r$ is upper bounded by $\frac{2}{2+c^2}$. The proof is in Appendix~\ref{sec:proof_4_1}
\end{theorem}

Under the null hypothesis, we assume that $f$ fails to capture $r$ for triplet $(u, v, v')$, and thus the embedding distance from $u$ to them is independent of their similarity w.r.t. $r$. The probability that their distance margin greater than or equal to $c\sigma_{D_f}$ is: 
\begin{align}\label{eq:random_cc_prob}
    \Prob\Big(d_f(u,v)-d_f(u,v')\geq c\sigma_{D_f} \: \Big\vert \: v,v'\in \V_t\Big)  
\end{align}

Theorem~\ref{thm:ub_cc_rate} gives the upper bound of the null hypothesis and the $c$ controls the strength of the test by moderating the size of the gap needed to satisfy clustering coherence. If clustering coherence rate is higher than this upper bound, our interpretation method will reject the null hypothesis and state that the node representation captures the node relation well.

\subsection{Smoothness Coherence}
\label{sec:rank_coherence}

Next, we develop {\em smoothness coherence} as a second approximation of weak relation coherence. Specifically, when considering a set of nodes that share high similarity w.r.t $r$ to query node $u$, the ranking of nodes in the set induced by their similarity in $r$ should be captured in the ranking induced by their distances in the embedding space. This coherence is important for the task that producing outcomes from nodes sharing high similarity, e.g. link prediction.

Formally, we define {\em similar nodes} for a query node $u$ as 
\begin{align*}
\mathcal{M}_r(u,\eta_1, \cdots ,\eta_k)=\{v_{s_i}  \: &| \: v_{s_i}\in C_{r,u}(\eta_i)/C_{r,u}(\eta_{i-1}) \notag\\
&\wedge v_{s_i} \!\in\! \V_t, i=1,\cdots,k\}, 
\end{align*}
which $\eta_i>\eta_s \forall i=1,\cdots,k$ and $\eta_{i-1}>\eta_i$. For notation simplicity, we define $\mathcal{M}_r(u,k)=\mathcal{M}_r(u,\eta_1, \cdots ,\eta_k)$ and $|\mathcal{M}_r(u,k)|=k$. Note that this $\mathcal{M}_r(u,k)$ is an ordered set satisfying $s_r(u, v_{s_1})>\cdots>s_r(u, v_{s_k})$. We set $\eta_i=\frac{k-i}{k}(1-\eta_s)+\eta_s$ and $\eta_{0}=1$. This corresponds to $\delta=\eta_{i-1}-\eta_i$ for weak coherence in Def.~\ref{def:weak_sc}. The incoherence for nodes with similarity value between $\eta_{i-1}$ and $\eta_i$ is ignored.

To express the rankings induced from different measures, we define two ranking functions to obtain orderings of the node set: the first, $\rho^{+}(u, q, \mathcal{M}_r(u,k))$, is a ranking function that ranks nodes from $\mathcal{M}_r(u,k)$ in {\em increasing} order w.r.t. node $u$, according to value function $q:\V\times\V\rightarrow \Real$; the second, $\rho^{-}(u, q, \mathcal{M}_r(u,k))$, is a similar ranking function, but it ranks nodes in {\em decreasing} order using $q$.

Then, we define the smoothness coherence rate as the probability that, given the rank ordering of $\rho^{-}(u, s_r, \mathcal{M}_r(u,k))$ using $s_r$ as the value function, we obtain the same rank ordering from $\rho^{+}(u, d_f, \mathcal{M}_r(u,k))$ using $d_f$ as the value function, which is defined as:


\begin{definition} \textbf{Smoothness Coherence Rate}.
For a model $f$, relation $r$, query node $u$, and 
set of nodes $\mathcal{M}_r(u,k)$ that are similar to $u$ due to relation $r$, the smoothness coherence rate corresponds to the probability:
\begin{small}
\begin{align}\label{eq:sc_prob}
    \Prob&\Big(\rho^{+}(u, d_f, \mathcal{M}_r(u,k)) \: \Big\vert \: \rho^{-}(u, s_r, \mathcal{M}_r(u,k)) \Big)=\notag\\
    \Prob&\Big(d_f(u, v_{s_1})<\cdots<d_f(u, v_{s_k})\: \Big\vert \: s_r(u, v_{s_1})>\cdots>s_r(u, v_{s_k})\Big)
\end{align}
\end{small}

To derive smoothness coherence rates, we define a function to calculate the empirical probability of Eq.~\eqref{eq:sc_prob} over a set of query nodes $\V_h$, as follows:

\begin{align}\label{eq:sc_test}
    &\Gamma_{smo}(f,r,\eta_s,k)=\notag\\
    &\frac{1}{|\V_h|}\sum_{u \in \V_h} \Ind\Big[\rho^{+}(u, d_f, \mathcal{M}_r(u,k))=\rho^{-}(u, s_r, \mathcal{M}_r(u,k))\Big],
\end{align}
\end{definition}
\noindent which estimates the level of smoothness coherence for $f$ due to $r$. The indicator function in Eq.~\eqref{eq:sc_test} considers a group of similar nodes $\mathcal{M}_r(u,k)$ as well as their ranks according to $\rho^{-}(u, s_r, \mathcal{M}_r(u,k))$, and yields 1 if $\rho^{+}(u, d_f, \mathcal{M}_r(u,k))$ returns the same ranking in reverse (due to the use of distance in embedding space instead of similarity). Hence, being an average over all nodes in the query set, this is an estimate of the probability in Eq.~\eqref{eq:sc_prob}.

However, just as in clustering coherence, the variance of the embedding distance distribution can impact the likelihood that node representation $Z$ preserves the smoothness coherence just by chance, instead of actual capturing the node relation. Recall that under the null hypothesis, we assume that $f$ fails to capture $r$. Then, the embedding distance from $u$ to $v_s \in \mathcal{M}_r(u,k))$ is independent of the similarity between $u$ and all $v_{s}$, and Eq.~\eqref{eq:sc_prob} becomes
\begin{align}\label{eq:random_sc_prob}
\Prob\left(d_f(u, v_{1})<\cdots<d_f(u, v_{k})\: \Big\vert \:  [v_{1}, \cdots, v_{k}] \in\V_t^k\right).
\end{align}
This is the probability that, if we iteratively sample $k$ nodes out of $\V_t$ at random, the first node will have smallest distance to $u$, the second node will have the second smallest until, consecutively, the $k$-th node will have the largest distance among them. Although we do not have a closed form analytical bound for Eq.~\eqref{eq:random_sc_prob} under the null hypothesis as we do for Eq.~\eqref{eq:random_cc_prob}, we can estimate the bound empirically and the details are in Appendix~\ref{sec:smooth_bound}.

\subsection{Interpretation Scores}
\label{sec:application}

Finally, we define {\em Coherence Rate} as our interpretation score of {\method}, which quantitatively assesses how faithfully a particular node relation $r$ is captured in the embedding space: 
\begin{definition}\textbf{Coherence Rate}. With the clustering coherence rate defined in Eq.~\eqref{eq:cc_test} and smoothness coherence rate  defined in Eq.~\eqref{eq:sc_test} for a node relation $r$, the quantitative measure of weak relation coherence is defined as:
\begin{equation}\label{eq:lcs}
    \mathbf{\Gamma}_{r,f} = \text{Agg}\left(\Gamma_{clu}(f,r,c,\eta_i,\eta_s,k), \Gamma_{smo}(f,r,\eta_s,k)\right).
\end{equation}
\end{definition}

\noindent Coherence rate summarizes weak relation coherence for model $f$ with values in the range $[0,1]$. A value of $1$ indicates that generated embeddings from $f$ are most expressive embeddings for $r$. We use the mean as the $Agg$ function in this paper. 

We then further define a {\em Model Coherence Score} to indicate which node representation learning approach encompass a larger number of node relations, to a higher degree:
\begin{definition}\textbf{Model Coherence Score}. 
\begin{equation}\label{eq:model_score}
    \Omega_f = \sum_{r\in\R}\omega_{r}\mathbf{\Gamma}_{r,f},
\end{equation}
\end{definition}
subject to $\sum_{r\in\R}\omega_{r}=1$. The weights can be decided by the machine learning practitioners, depending on their relations of interest. The model coherence score is derived from the coherence rates of {\method}. In Section~\ref{sec:model_selection}, we demonstrate the strong linear correlation between $\Omega_f$ and the performance on node classification and link prediction.

\begin{small}
\begin{table}[t]
\caption{Data Statistics.}
\vspace{-4mm}
\label{tab:dataset}
\begin{center}
\begin{tabular}{lrrrr}
\hline
Dataset & Node & Edge & Node Class & Node Attr. \\
\hline
Cora & 2708 & 10556 & 7 & 1433 \\
CiteSeer & 3327 & 9104 & 6 & 3703 \\
Brazil & 131 & 1074 & 4 & - \\
USA & 1190 & 13599 & 4 & - \\
Computers & 13752 & 491722 & 10 & 767 \\
Photo & 7650 & 238162 & 8 & 745 \\
\hline
\end{tabular}
\vspace{-4mm}
\end{center}
\end{table}
\end{small}

\section{Experiments}
We conducted experiments to address the following research questions: {\bf RQ1:} How do different interpretation methods compare in terms of their interpretation accuracy using our {\eval}? and {\bf RQ2:} What can we learn from the application of {\method} interpretation? For RQ2, the interpretation results can be used to answer following subquestions: {\bf RQ2-1:} What node relations are captured by the node representations generated from different graph-based methods? {\bf RQ2-2:}  How does the performance on downstream tasks depend on the captured node relations? and {\bf RQ2-3:} Can we choose models based on coherence rate in an unsupervised setting?

\vspace{0.5em}
\noindent\textbf{Dataset.}  The data statistics for six graphs are in Table~\ref{tab:dataset} and the detailed description is in Appendix~\ref{sec:dataset}.

\vspace{0.5em}
\noindent\textbf{Graph-based Models.} We choose five graph-based models to generate node representations for node representation interpretation. We have three GNN-based models (GCN~\cite{gcn}, GAT~\cite{gat}, and DGCN~\cite{dgcn}), one matrix projection model (FastRP~\cite{fastrp}) and one skip-gram model (Node2Vec~\cite{node2vec}). The detailed model description is in Appendix~\ref{sec:graph_model}.

\vspace{0.5em}
\noindent\textbf{Compared Interpretation Methods.} There are two main types of interpretation methods. The first group of interpretation methods use a Kendall's $\tau$ ranking test~\cite{kandell, eval_stuct}. This approach involves evaluating the degree of relation coherence by comparing the similarity of each node relation to the embedding distance. The other type of approach is Property Classification~\cite{intrinsic_struct_eval,eval_embed1,eval_embed2,node_embed_classification}, which formulates representation interpretation as an intrinsic property classification task. This groups of methods is referred as "Class" in Table~\ref{tab:interpret_mrr}. The experiment settings of {\method} is described in Appendix~\ref{sec:test_exp_setup}.


\begin{figure*}[t]
  \centering
  \includegraphics[width=\linewidth]{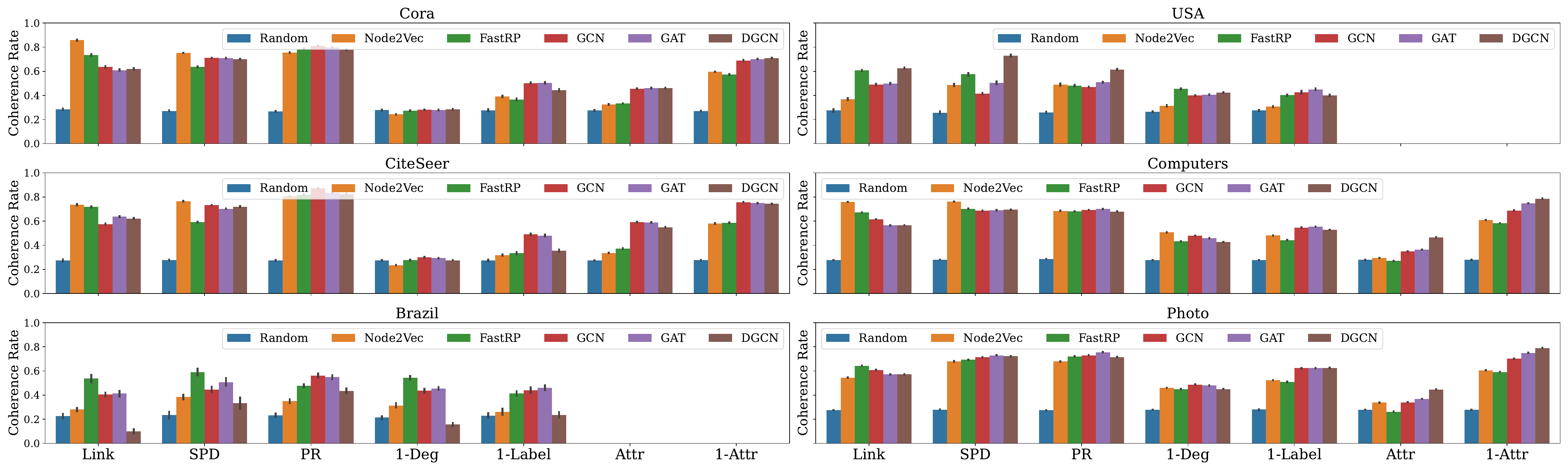}
  \caption{Interpreting Node Representations with {\method}. A high coherence rate of a node relation suggests that the model captures the node relation well in its embedding space.}
  \label{fig:coherence_test}
  \vspace{-3mm}
\end{figure*}

\begin{table}[t]
\caption{Interpretation accuracy (MRR). The previous approaches have low MRR because either their interpretations are affected by expressiveness of embeddings or critical incoherence.}
\vspace{-4mm}
\label{tab:interpret_mrr}
\setlength{\tabcolsep}{2pt}
\begin{center}
\begin{tabular}{lccccccc}
\hline
Method & Cora & CiteSeer & Brazil & USA & Photo & Computers \\
\hline
Kendall's $\tau$ & 0.89 & 0.78 & 0.61 & 0.70 & 0.81 & 0.88 \\
Class & 0.38 & 0.44 & 0.45 & 0.50 & 0.56 & 0.56 \\
\hline
{\method}& 0.93 & 0.83 & 0.67 & 0.73 & 0.90 & 1.00 \\
\hline
\end{tabular}
\vspace{-4mm}
\end{center}
\end{table}

\begin{figure}[h]
  \centering
  \includegraphics[width=0.7\linewidth]{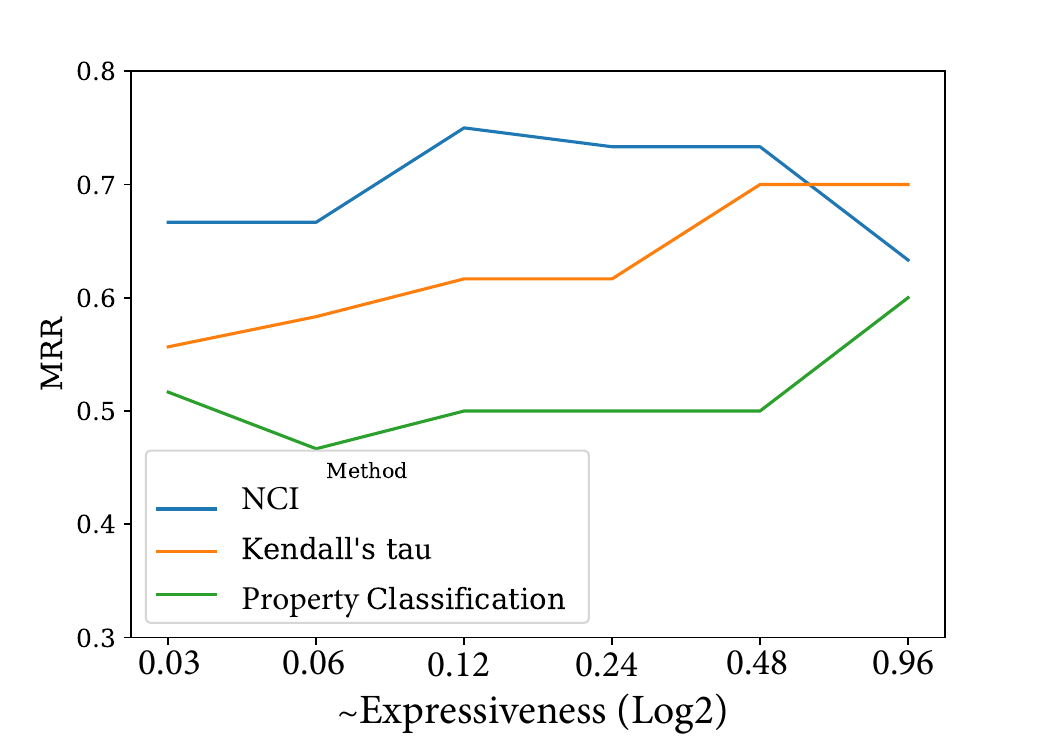}
  \caption{When the embedding is less expressive, our {\method} has better interpretation accuracy (MRR) compared to other representation interpretation methods.}
  \label{fig:svd_dim}
  \vspace{-9mm}
\end{figure}

\subsection{Node Relations}
\label{sec:node_relation}
The node relations we consider are often used to describe general graph characteristics. They are illustrative examples for our framework and not intended to be the definitive set of relations for all graph domains. Note that if, in a particular application, other relations are of interest, those relations can be seamlessly integrated into the {\method} measures. We use training data to calculate the similarity values for Node Coherence Test. Below we specify the node similarity function for each node relation and note that similarity only needs to be calculated once for each graph.

\vspace{1mm}
\noindent\textbf{Has Link (Link).}~\cite{gcn} 
\begin{equation}
    s_{\text{link}}(u,v)=\widetilde{D}^{-\frac{1}{2}}\widetilde{A}\widetilde{D}^{-\frac{1}{2}}[u,v],
\end{equation}
where $\widetilde{A}=A+I$ is the adjacency matrix $A$ with self-loop and $\widetilde{D}_{uu}=\sum_{v}\widetilde{A}_{uv}$ and $[u,v]$ indicates the $(u,v)$-entry of $\widetilde{D}^{-\frac{1}{2}}\widetilde{A}\widetilde{D}^{-\frac{1}{2}}.$ This similarity is inspired by \cite{gcn} to weight each link.

\vspace{1mm}
\noindent\textbf{Shortest Path Distance (SPD).}~\cite{shortest_path} 
\begin{equation}
    s_{\text{shortest-path}}(u,v)=\frac{\text{diameter}(\G)-d_{G}(u,v)+1}{\text{diameter}(\G)},
\end{equation}
where $d_G(u,v)$ is the shortest-path distance between $u$ and $v$. Note that $s_{\text{shortest-path}}(u,v)=\text{diameter}(\G)+1$ when $u$ and $v$ are disconnected in a graph, and hence $s_{\text{shortest-path}}(u,v)=0$. 

\vspace{1mm}
\noindent\textbf{PageRank (PR).}~\cite{pagerank}  We can derive $\pi_u$ of personalized PageRank by solving the following equation
\begin{equation*}
    \pi_u=\alpha P\pi_u + (1-\alpha)e_u,
\end{equation*}
where $P$ is the transition probability and $e_u\in[0,1]^{|\V|}$ is a one-hot vector containing a one at the position corresponding $u$'s entry and zeros otherwise. The similarity in terms of PageRank is 
\begin{equation}
    s_{\text{PageRank}}(u,v)=\pi_{u,v}.
\end{equation}

\vspace{1mm}
\noindent\textbf{Degree Distribution. (1-Deg)}~\cite{struct2vec}  
\begin{equation}
    s^k_{\text{degree distribution}(u,v)} = \cos(\text{deg}^k_u,\text{deg}^k_v),
\end{equation}
where $\text{deg}^k_u\in[0, 1]^{\delta(\G)}$ is the $k$-hop neighbor degree distribution of $u$ and $\delta(\G)$ is the maximum degree in graph $\G$. We set $k=1$ in our experiments.

\vspace{1mm}
\noindent\textbf{Label Distribution. (1-Label)}~\cite{lp} 
\begin{equation}
    s^k_{\text{attribute distribution}}=\cos(y^k_u, y^k_v),
\end{equation}
where $y^k_u=Normalize(A^kY)_u$ is the $k$-hop neighbor attribute distribution and $C$ can be the dimension of the number of categories. We use $k=1$ and 1-Label is the class label distribution of $1$-hop neighbors in our experiments.

\vspace{1mm}
\noindent\textbf{Attribute Similarity. (Attr)} 
\begin{equation}
    s^k_{\text{attribute similarity}}=\cos(x_u, x_v),
\end{equation}
where $x_u\in\Real^{D}$ is the attributes of node $u$ and $D$ is the dimension.

\vspace{1mm}
\noindent\textbf{Attribute Distribution. (1-Attr)}~\cite{lp} 
\begin{equation}
    s^k_{\text{attribute distribution}}=\cos(x^k_u, x^k_v),
\end{equation}
where $x^k_u=Normalize(A^kX)_u$ is the $k$-hop neighbor attribute distribution. We use $k=1$ and 1-Attr is the node attribute distribution of $1$-hop neighbors in our experiments.

\subsection{RQ1: Accuracy of Interpretation Methods}
\label{sec:exp_interpret_acc}

We compare three types of interpretation methods  and show that our {\method} 
outperforms the others. We evaluate each interpretation method using the {\eval} described in Section~\ref{sec:eval_process}. The embedding dimension $d$ is set to $256$ for Brazil and $512$ for the rest of the dataset. The node relation set $\R$ includes all the node relations defined in Section~\ref{sec:node_relation}. The performance metric is Mean-Reciprocal-Rank (MMR). The higher the MMR, the better interpretation accuracy. 

According to MRR in Table~\ref{tab:interpret_mrr}, our {\method} 
has higher interpretation accuracy compared to other methods. Formulating interpretation as an intrinsic property classification (ie. Class) emphasizes too much on the cluster property yet neglects critical incoherence in high similarity regimes, specifically smoothness coherence. Consequently, they fail to identify the node relations generating the embeddings by EVD. Additionally, the Kendall's $\tau$ ranking test, which interprets embeddings based on relation coherence, often performs poorly because in practical settings when the expressivity of the representations are lossy the interpretation accuracy are degraded.
Table~\ref{tab:interpret_mrr} empirically shows that our {\method} has better interpretation accuracy because it's based on weak relation coherence which is less affected by the expressiveness of embeddings.

{\bf Embedding Expressiveness vs Interpretation Accuracy}
Next, we varied the expressiveness of node representation $Z_{r_t}$ from $0.03$ (least expressive) to $0.96$ (close to most expressive) to assess the impact of noise on interpretation accuracy, as shown in Figure~\ref{fig:svd_dim}. We define expressiveness as $d/|\V|$, which is an empirical approximation of expressiveness based on the discussion in Section~\ref{sec:eval_process}. (Note this does not apply for a more general ML setting.) We use the USA dataset for this experiment. When expressiveness is close to $1$, Kendall's $\tau$ is expected to yield  accurate results. This is evident in Figure~\ref{fig:svd_dim}.  However, when the expressiveness is less than $0.5$, there is a significant amount of incoherence. Some triplets with a small distance difference confuse the interpretation from Kendall's $\tau$ with the node relations which similarity values concentrate in a narrow range. This finding supports our argument that, {\em when the embeddings are less expressive, methods based on weak relation coherence will be more accurate than approaches adhering to full relation coherence}.

\vspace{-2mm}
\subsection{RQ2-1:Node Representation Interpretation}
\label{sec:coherence_test}

We use {\method} to analyze the node representations from five graph-based models for six graphs in Figure~\ref{fig:coherence_test}. Random represents the average upper bounds of coherence rates under the null hypothesis for all graph-based models, as explained in Section~\ref{sec:framework}. Coherence rates of any graph-based models should be greater than coherence rates of Random to indicates the relations are captured by the model significantly. All models are trained in an unsupervised setting to initially produce node representations. We exclude the results of Attribute Similarity and 1-hop Attribute Distribution for USA and Brazil because these graphs don't have node attributes. We demonstrate the ability of each model to capture seven node relations in Section~\ref{sec:node_relation}. The details of experiment settings for {\method} in Appendix~\ref{sec:test_exp_setup}.


\vspace{1mm}
{\bf Link}: Link relation is a structural relation. Node2Vec has better capability to capture Link relation in homophilious graphs, in which nodes having similar properties are usually connected~\cite{homo1} (Cora, CiteSeer, Computers and Photos), but FastRP is better than Node2Vec in non-homophilious graphs~\cite{homo2} (USA, Brazil).

\vspace{1mm}
{\bf SPD}: SPD relation is another structural relation. Node2Vec has better capability to capture SPD relation in most of homophilious graphs (Cora, CiteSeer, and Computers). FastRP is good at capturing SPD relation for non-homophilious graphs, but DGCN has better performance in USA because it has better ability to capture higher-order relation~\cite{dgcn}.

\vspace{1mm}
{\bf PR}: PR relation is a proximity-based relation. Nodes within local subgraph usually have higher similarity between them. All models have comparable ability to capture PR relation in most of graphs~\cite{homo1} because their learning algorithms usually derives information from a local subgraph given a node.

\vspace{1mm}
{\bf 1-Deg}: 1-Deg relation is a structural relation. All models cannot capture 1-Deg relation in Cora and CiteSeer (they all have comparable or lower coherence rates than Random) because it's difficult for them to capture 1-Deg relation in a sparse graph, in which nodes usually have low node degrees.

\vspace{1mm}
{\bf 1-Label, Attr, 1-Attr}: They are all attribute-based relations. These node attributes or labels are used to describe the properties of a node and nodes with similar attributes usually belongs to the same local subgraphs. GNN models (GCN, GAT, and DGCN) are good at capturing these relations in most of the graph because their model architectures usually help them to learn local community structures.


\vspace{1mm}
The summary from our interpretation results is that: (1) GNN-based models are usually good at capturing attribute-based relations, e.g. 1-hop Label Distribution or 1-hop Attribute Distribution, (2) Node2Vec or FastRP can be used if structural information is important for a downstream task, and (3) none of the graph-based methods can capture 1-hop Degree Distribution well for Cora and CiteSeer.

\subsection{RQ2-2: Correlation between Task Performance and Coherence Rate}

We apply Pearson Correlation Coefficient analysis to measure the linear correlation between coherence rates of models for each node relation and their performance on node classification and link prediction. We compute the correlation coefficients between the performance of five graph-based models on each task and their coherence rates of each node relation. Table~\ref{tab:link_corr} shows the correlation analysis performance of link prediction and coherence rate for each node relation. Table~\ref{tab:nc_corr} shows the correlation analysis between performance of node classification and coherence rate for each node relation. The results in Table~\ref{tab:link_corr} and Table~\ref{tab:nc_corr} show the importance of different node relations on the success of node classification and link prediction. The higher the coefficient value, the more important a node relation is. For example, The labels in the USA and Brazil indicates the activity level of a airport, which highly relates to the degree of a node (airport). Therefore, the coefficients of 1-Deg are higher compared to other datasets. The experiment setting for node classification and link prediction is discussed in Appendix~\ref{sec:test_exp_setup}.

\begin{table}[t]
\caption{Correlation Link Prediction and Node Relation.}
\setlength{\tabcolsep}{2pt}
\label{tab:link_corr}
\begin{center}
\begin{tabular}{lcccccccc}
\hline
Dataset & Link & SPD & PR & 1-Deg & 1-Label & Attr & 1-Attr \\
\hline
Cora & -0.63 & 0.98 & 0.95 & 0.67 & 0.97 & 0.95 & 0.98 \\
CiteSeer & -0.14 & 0.99 & 1.00 & 0.76 & 0.88 & 1.00 & 1.00 \\
Brazil & -0.19 & 0.94 & 0.85 & 0.27 & 0.46 & - & - \\
USA & -0.99 & 0.81 & 0.65 & -0.63 & -0.02 & - & - \\
Computers & 0.90 & 0.73 & 0.70 & 0.90 & -0.51 & -0.91 & -0.93 \\
Photo & 0.99 & 0.61 & 0.98 & 0.48 & 0.40 & -0.17 & 0.18 \\
\hline
\end{tabular}
\end{center}
\end{table}

\begin{table}[t]
\caption{Correlation Node Classification and Node Relation.}
\setlength{\tabcolsep}{2pt}
\label{tab:nc_corr}
\begin{center}
\begin{tabular}{lcccccccc}
\hline
Dataset & Link & SPD & PR & 1-Deg & 1-Label & Attr & 1-Attr \\
\hline
Cora & -0.36 & 0.79 & 0.20 & 0.07 & 0.82 & 0.66 & 0.76 \\
CiteSeer & -0.85 & 0.62 & 0.58 & 0.47 & 0.69 & 0.89 & 0.94 \\
Brazil & 0.74 & 0.67 & 0.86 & 0.78 & 0.97 & - & - \\
USA & 0.86 & 0.45 & 0.46 & 0.88 & 0.86 & - & - \\
Computers & 0.44 & 0.21 & 0.58 & 0.60 & -0.14 & -0.82 & -0.60 \\
Photo & 0.65 & 0.66 & 0.88 & 0.52 & 0.52 & -0.12 & 0.36 \\
\hline
\end{tabular}
\end{center}
\end{table}

\subsection{RQ2-3: Model Selection}
\label{sec:model_selection}
\begin{figure}[t]
\centering
\begin{subfigure}{.5\linewidth}
  \centering
  \includegraphics[trim={0 0 0 0},clip,width=\linewidth]{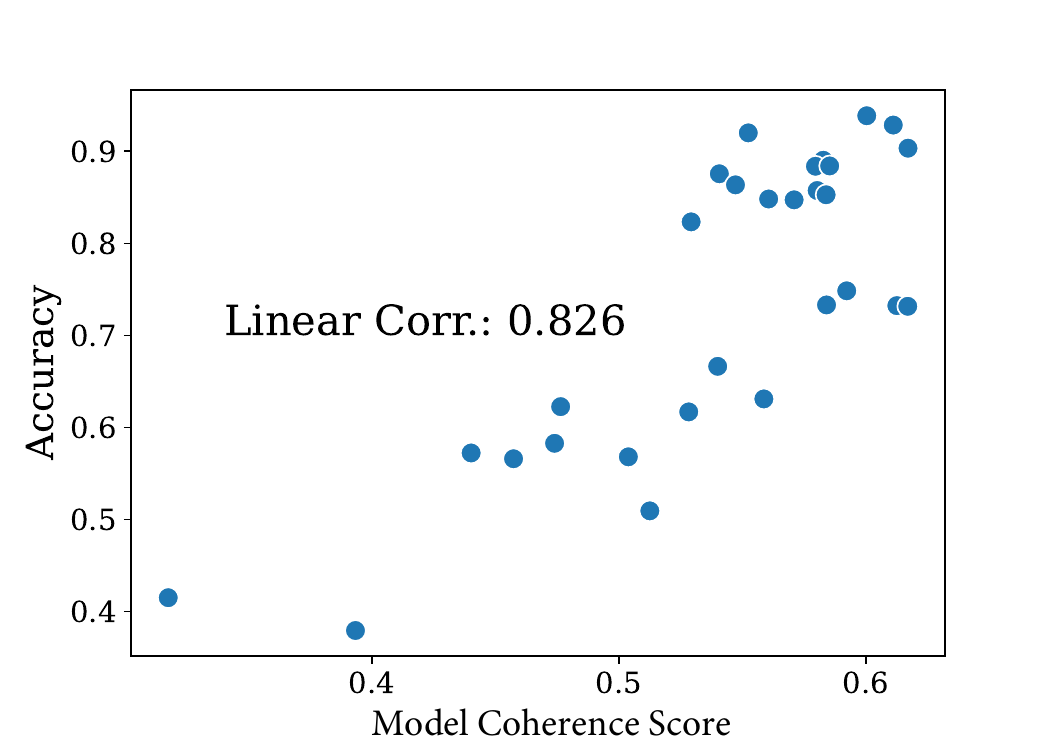}
  \caption{Node Classification.}
  \label{fig:nc_model_select}
\end{subfigure}%
\begin{subfigure}{.5\linewidth}
  \centering
  \includegraphics[trim={0 0 0 0},clip,width=\linewidth]{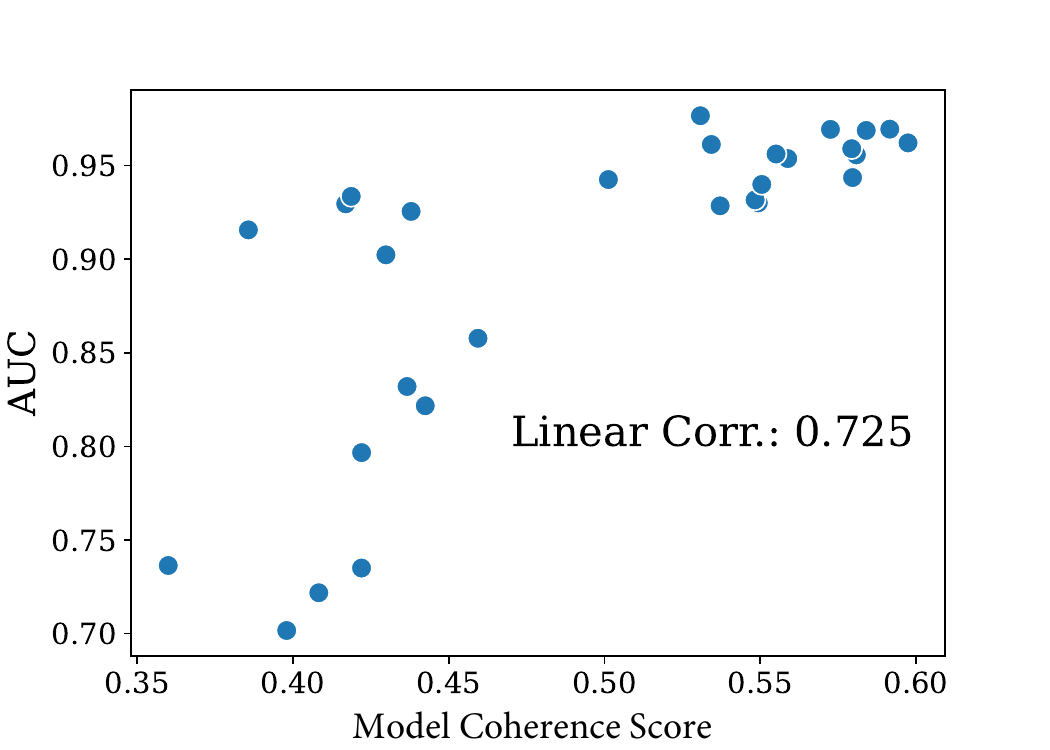}
  \caption{Link Prediction.}
  \label{fig:link_model_select}
\end{subfigure}
\caption{Model Coherence Score for Model Selection in unsupervised settings. When a model coherence score of a model is higher, the model tends to have better performance on downstream tasks.}
\label{fig:model_select}
\vspace{-2mm}
\end{figure}

In this section, we calculate the Pearson Correlation Coefficient between our model coherence score for node representations in Eq.\eqref{eq:model_score} and performance metrics on downstream task. Each dot in Figure~\ref{fig:model_select} shows a model's performance on a downstream task and its model coherence score in one graph. High positive correlation coefficients indicate that the model coherence score can be used for model selection, even in an unsupervised setting without prior knowledge of the downstream task. Figure~\ref{fig:model_select} shows that our model coherence score are highly correlated with the performance on both node classification and link prediction. It means that the performance on the task is better as the model coherence score is higher. Therefore, we empirically demonstrate that our  model coherence score can be used to select models in unsupervised settings.

\section{Related Work}

The field of Explainable AI (XAI) has emerged to enhance the interpretability of models and provide explanations for model decisions.~\cite{myth_interpretability,xai,interpret_explain_dnn}. Numerous techniques for XAI for Deep Neural Network have been proposed, ranging from algorithm transparency~\cite{polar,decesion_tree}, global model interpretability~\cite{pgexplainer}, local interpretability~\cite{lime,lpr} to representation interpretation~\cite{word_intrusion_test,tobias,qvec}.

XAI for Graph Neural Networks (GNNs) has become popular as they have shown exceptional performance in various graph-related applications~\cite{lpr,graphlime,xgnn}. However, XAI for GNN only focuses on providing the features and parameters behind a model decision~\cite{sa,gradcam,gnnexplainer,graphmask, gnn_explain}, They overlook making the output node representation understandable for humans, which is interpreting node representation. The absence of node representation interpretation poses challenges to the model's reliability and the decisions made by the task-specific model.

There is limited literature on interpretation techniques for node representations. \cite{eval_stuct, n_proximity} employed statistical correlation tests to assess the accuracy of encoding structural patterns in node representations. Conversely, there is another cluster of studies~\cite{intrinsic_struct_eval,eval_embed1,eval_embed2,node_embed_classification} that approach interpretation as a classification task. These investigations used classification accuracy as  interpretation scores to assess a model's ability to accurately capture diverse relations within the embedding space. However, the accuracy of their interpretation results is unknown due to a lack of quantitative evaluation algorithms in past work. As we show  in Section~\ref{sec:exp_interpret_acc}, these two methods have lower accuracy in interpreting node embeddings.

\vspace{-2mm}
\section{Conclusion}

In this paper, we introduced a novel method for node representation interpretation---{\fullmethod} ({\method})---which measures the extent to which node representations capture the node relations in the original graphs. We also proposed  a novel Interpretation Method Evaluation (IME) Process to validate the reliability of various interpretation methods. We applied {\method} to interpret the various node representations generated by different graph-based models and analyzed the quality of the learned node representations. {\method}, based on weak relation coherence, has higher interpretation accuracy compared to other interpretation methods when the embeddings are less expressive. Furthermore, our interpretation scores, referred to as coherence rates, show that our {\method} can offer valuable insights into the strengths and weaknesses of various node representations. Additionally, the connection between coherence rates and performance on downstream tasks can uncover the crucial factors that influence performance on those tasks. Finally, we show that {\method} model coherence scores can be used effectively for model selection in unsupervised settings.

\bibliographystyle{ACM-Reference-Format}
\bibliography{sample-base}

\appendix
\section{Proof of Lemma 3.1}
\label{sec:proof_3_1}
\begin{proof}
Given a triplet $(u,v,v')$ and $s_r(u,v)>s_r(u,v')$
    \begin{align*}
        &s_r(u,v)>s_r(u,v')\rightarrow s_r(u,v)-s_r(u,v')>0\\
        \rightarrow&z_uz_v^{\top}-z_u(z'_v)^{\top}>0\\
        \rightarrow&\frac{\|z_u\|^2+\|z_v\|^2-d_f(u,v)^2}{2}-\frac{\|z_u\|^2+\|z'_v\|^2-d_f(u,v')^2}{2}>0 \\
        \rightarrow&\|z_v\|^2-\|z'_v\|^2+d_f(u,v')^2-d_f(u,v)^2>0 \\
        \rightarrow&d_f(u,v')^2-d_f(u,v)^2>0\rightarrow d_f(u,v)<d_f(u,v).
    \end{align*}
From vector calculation between any two vectors $x$ and $y$, the relation $(\|x-y\|^2=\|x\|^2+\|y\|-2xy)$ exists and $\|z_v\|^2=1$ because $S_r=Z_rZ_r^{\top}.$ Similar proving process can be applied to proof that if $s_r(u,v)=s_r(u,v')$, then $d_f(u,v)=d_f(u,v)$
\end{proof}




\section{Proof of Theorem 4.1}
\label{sec:proof_4_1}
\begin{proof}
To proof this, we use Cantelli's inequality\cite{cantelli} to bound the probability of distinguishing similar node cluster $C_{r,u}(\eta_s)$ and intruder cluster $C'_{r,u}(\eta_i)$ by random chance. Cantelli's inequality is an improved version of Chebyshev's inequality for one-sided bounds, which states:
\begin{equation*}
    \Prob(X-\Exp[X]\geq \epsilon) \leq \frac{\sigma_{X}^2}{\sigma_{X}^2+\epsilon^2},
\end{equation*}
where $X$ is a real-valued random variable and $\epsilon>0$ is a constant.

For our clustering coherence scenario, let $X: d_f(u,v)-d_f(u,v')$ and $\epsilon=c\sigma_{D_f}$. Under the null hypothesis, we will assume that $f$ fails to capture $r$ for triplet $(u, v, v')$, and thus the embedding distance from $u$ to them is independent of their similarity w.r.t. $r$. The probability that their distance margin greater than or equal to $c\sigma_{D_f}$ is: 
\begin{align}
    \Prob\left(d_f(u,v)-d_f(u,v')\geq c\sigma_{D_f} \: \Bigg\vert \: v,v'\in \V_t\right)  
\end{align}

Moreover, as $d_f(u, v)$ and $d_f(u, v')$ are i.i.d. sampled from $D_f$ and independent of any relation, $\Exp[X]=0$ and $\sigma^2_X=2\sigma_{D_f}^2$ due to the independence between $d_f(u,v)$ and $d_f(u,v')$. Applying Cantelli's inequality, the upper bound is
\begin{align}\label{eq:ub}
    \Prob\left(d_f(u,v_i)-d_f(u,v_s)\geq c\sigma_{D_f}\right) &\leq \frac{2\sigma_{D_f}^2}{2\sigma_{D_f}^2+(c\sigma_{D_f})^2}\notag \\
    &=\frac{2}{2+c^2}.
\end{align}
The above equation states that the upper bound of \eqref{eq:random_cc_prob} is a constant $\frac{2}{2+c^2}$, regardless of graph-based model $f$. 
\end{proof}

\begin{figure*}[t]
  \centering
  \includegraphics[width=\linewidth]{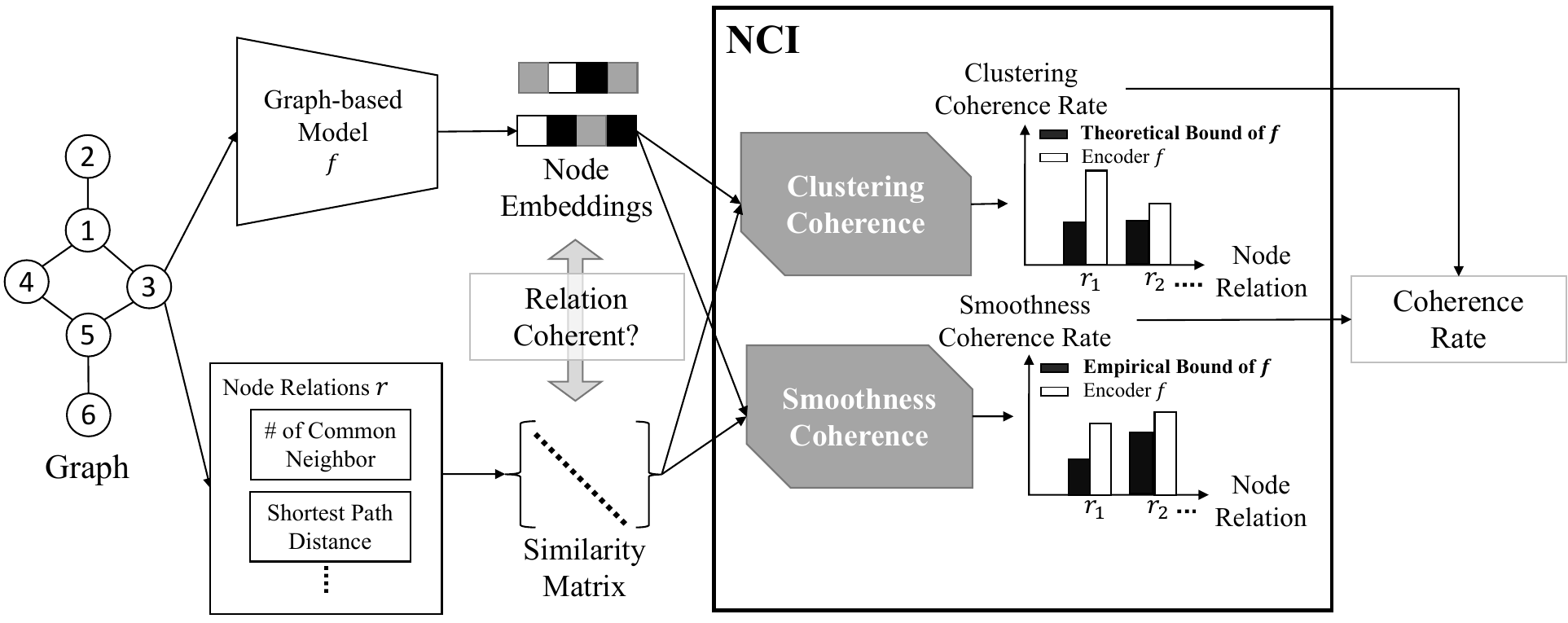}
  \caption{Interpretation pipeline for NCI.}
  \vspace{4mm}
  \label{fig:framework}
\end{figure*}

\section{Empirical Upper Bound for Smoothness Coherence Rate}
\label{sec:smooth_bound}
Given a node embedding for $f$, we randomly permute the mapping of nodes to embedding locations in order to generate a shuffled random embedding $f'$. This ensures that $f'$ has the same distance distribution $D_f$, but that the node embeddings are independent of any $r\in\R$. We repeat this process several times to generate a set of random node embeddings $\mathbf{F}'$. Then, we derive smoothness coherence rate as Eq.(4) to all random embeddings $f' \in \mathbf{F}'$ and derive the empirical distribution of $\overline{\Prob}(d_f(u, v_{s_1})< \cdots <d_f(u, v_{s_k}))$. Since, by the central limit theorem, $\overline{\Prob}$ is normally distributed,  we can calculate its empirical mean and standard deviation and set the upper bound of Eq.(5) as the $95$th-percentile of the distribution. Similar as clustering coherence rate, if the smoothness coherence rate is greater than this upper bound, we will state that the semantic relation is well captured in the embedding space.

\section{Datasets}
\label{sec:dataset}

Table~\ref{tab:dataset} shows the basic statistics of each graph.
\begin{itemize}
    \item Cora and CiteSeer~\cite{cora} are citation networks. Each node represents a documents and citation links exist between documents. Each document belongs to one research area, which is the label on each node. 
    \item USA and Brazil~\cite{struct2vec} are air traffic networks which were collected from the Bureau of Transportation Statistics\footnote{ https://transtats.bts.gov/}. It is an undirected graph which nodes are airports and edges are routes. Their class labels on nodes describe the activity level of each airport.
    \item Computers and Photo~\cite{amazon} are subgraphs of the Amazon co-purchase graph. Each product is represented by a node, and each link between two nodes means that those products are often bought together by customers. Each product also has a class label, which indicates the product category it belongs to.
\end{itemize}

\section{Graph-based Models}
\label{sec:graph_model}
\begin{itemize}
    \item Random pass rates are the average random pass rates of all encoders. The method we use to calculate the random pass rates for each model are the same as the process we discussed in Section 4.1 and Section 4.2 in our main content.
    \item GCN~\cite{gae} is an unsupervised learning algorithm for graph-structured data. It learns to re-construct graphs based on the Variational Auto-Encoder. It uses Graph Convolutional Networks~\cite{gcn}(GCN) as an encoder, where the embeddings approximate the first-order spectral graph convolutions. 
    \item GAT~\cite{gat} is similar to GCN, but introduces a self-attention layer to address the shortcomings of graph convolution.
    \item FastRP~\cite{fastrp} is a scalable and optimization-free algorithm for learning network embeddings. It constructs a node similarity matrix by encoding transitive relationships and uses sparse random projection on the matrix for dimension reduction. 
    \item Node2Vec~\cite{node2vec} is an algorithm to generate low-dimension feature vectors for graphs. The features are generated via maximizing the likelihood of preserving neighborhood. It also uses a flexible notion of neighborhood via biased random walks, which is key for richer node representation.
    \item DGCN~\cite{dgcn} is a state-of-the-art encoder for homogeneous graphs. Most types of graph neural networks suffer from over-smoothing, where the performance degrades when the number of layers increases. DGCN solves the over-smoothing problem by detaching the weight matrix from feature propagation to preserve expressive power. We adopt a similar learning algorithm as GAE and change the encoder to DGCN.
\end{itemize}

\section{Experiment Setting}
\label{sec:test_exp_setup}

\subsection{Node Coherence Rate for Representation Interpretation}
The parameter settings are as follows: We set $k=3$ for both the clustering coherence rate and the smoothness coherence rate and $c=1.64$ for all models. We analyze the distribution of similarity values of $r$ for query node $u$ and set the similarity threshold $\eta_s$ to the $70$th percentile of that distribution. Additionally, we set the intruder threshold $\eta_i$ to the $5$th percentile. To calculate the upper bound of smoothness coherence rates under null hypothesis, we create $100$ random embeddings for each $f$ and determine the upper bound for smoothness coherence rate under null hypothesis. 

\begin{figure*}[t]
  \centering
  \includegraphics[width=\linewidth]{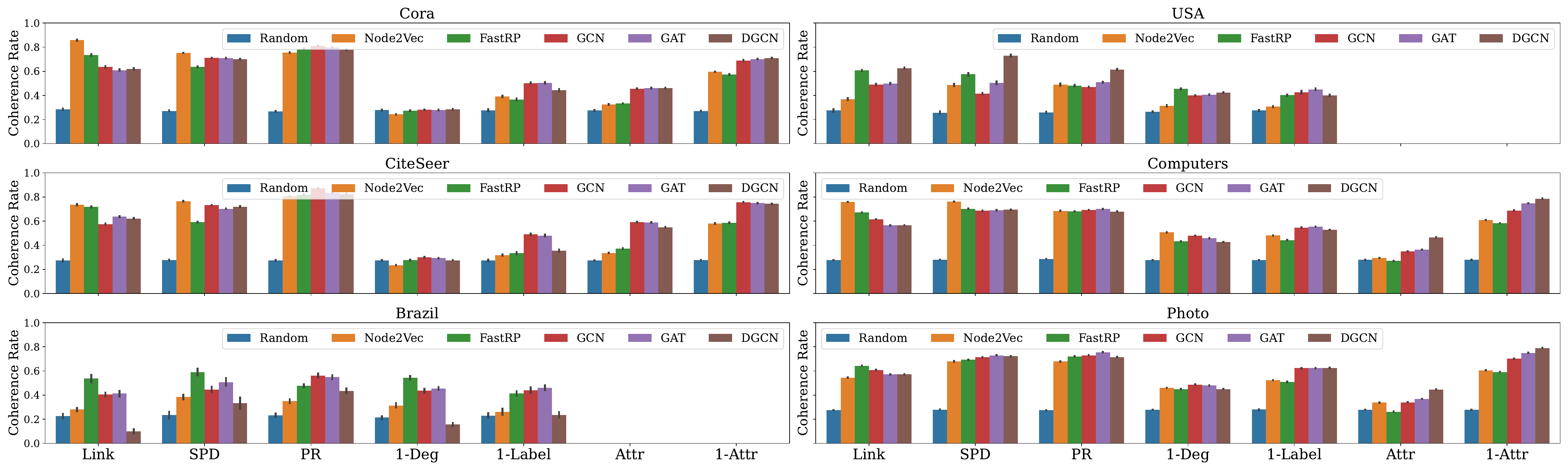}
  \caption{This figure shows the results of interpreting node representations. These node representations are generated based on the data setting for node classification. A high coherence rate of a semantic relation suggests that the model preserves the semantic relation well in the embedding space.}
  \label{fig:nc_coherence}
\end{figure*}

\begin{figure*}[t]
  \centering
  \includegraphics[width=\linewidth]{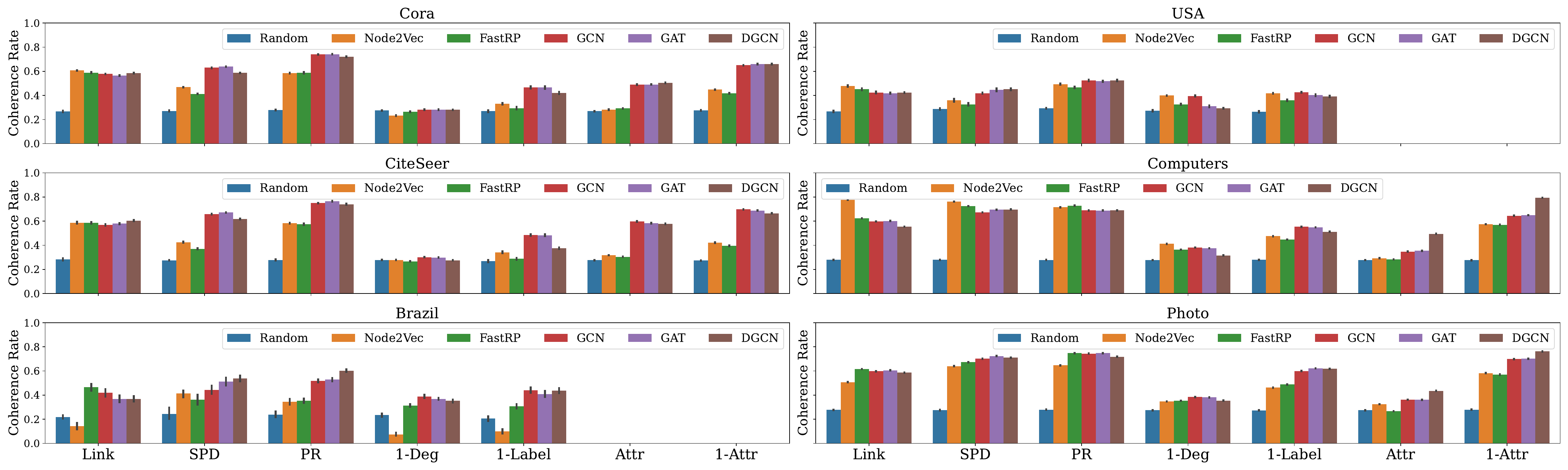}
  \caption{This figure shows the results of interpreting node representations. These node representations are generated based on the data setting for link prediction. A high coherence rate of a semantic relation suggests that the model preserves the semantic relation well in the embedding space.}
  \label{fig:link_coherence}
\end{figure*}

\begin{table}[t]
\caption{Performance on Node Classification (Accuracy).}
\vspace{-4mm}
\setlength{\tabcolsep}{2pt}
\label{tab:nc_acc}
\begin{center}
\begin{tabular}{lcccccc}
\hline
Method & Cora & CiteSeer & Brazil & USA & Computers & Photo \\
\hline
FastRP & 0.82 & 0.62 & 0.51 & 0.57 & 0.88 & 0.92 \\
GAE & 0.85 & 0.73 & 0.57 & 0.57 & 0.88 & 0.94 \\
Node2Vec & 0.85 & 0.67 & 0.42 & 0.38 & 0.88 & 0.86 \\
GAT & 0.86 & 0.73 & 0.62 & 0.58 & 0.89 & 0.93 \\
DGCN & 0.85 & 0.73 & 0.38 & 0.63 & 0.75 & 0.90 \\
\hline
\end{tabular}
\end{center}
\end{table}

\begin{table}[t]
\caption{Performance on Link Prediction (AUC).}
\vspace{-4mm}
\setlength{\tabcolsep}{2pt}
\label{tab:link_auc}
\begin{center}
\begin{tabular}{lcccccc}
\hline
Method & Cora & CiteSeer & Brazil & USA & Computers & Photo \\
\hline
FastRP & 0.72 & 0.70 & 0.74 & 0.92 & 0.96 & 0.98 \\
GAE & 0.93 & 0.96 & 0.82 & 0.93 & 0.96 & 0.97 \\
Node2Vec & 0.80 & 0.74 & 0.80 & 0.90 & 0.97 & 0.94 \\
GAT & 0.93 & 0.96 & 0.83 & 0.93 & 0.95 & 0.97 \\
DGCN & 0.93 & 0.94 & 0.86 & 0.93 & 0.94 & 0.96 \\
\hline
\end{tabular}
\end{center}
\end{table}

\subsection{Train/Test Split for Downstream Tasks}
For training a node classification model, we use $60\%$ of labels for training and $40\%$ for testing and all links in the graph for both training and testing. For link prediction, we use all labels for training and testing, and $70\%$ of links for training and $30\%$ of links for testing.

\end{document}